\documentclass[runningheads]{llncs}

\usepackage{eccv}

\usepackage{eccvabbrv}

\usepackage{graphicx}
\usepackage{booktabs}
\usepackage{multirow}
\usepackage{amssymb}
\usepackage{pifont}
\usepackage[dvipsnames]{xcolor}
\usepackage{tcolorbox}
\tcbuselibrary{breakable}
\usepackage{listings}
\usepackage{makecell}
\usepackage{array}
\usepackage{xspace}
\usepackage{wrapfig}
\usepackage{enumitem}

\usepackage[accsupp]{axessibility}  

 \usepackage{hyperref}

\usepackage{orcidlink}
\newcolumntype{R}[1]{>{\raggedleft\arraybackslash}p{#1}}
\newcommand{\name}{\textit{ATP-Bench}\xspace}

\newcommand{\cmark}{{\color{ForestGreen}\ding{51}}} 
\newcommand{\xmark}{{\color{Maroon}\ding{55}}}     
\definecolor{toolblue}{RGB}{31,119,180}

\makeatletter
\newcommand\blfootnote[1]{%
  \begingroup
  \renewcommand\thefootnote{}%
  \NoHyper\footnotetext{#1}\endNoHyper
  \endgroup
}
\makeatother

\begin{document}

\title{\name: Towards Agentic Tool Planning for MLLM Interleaved Generation} 

\titlerunning{\name}

\author{
Yinuo Liu\inst{1,2}\textsuperscript{*} \and
Zi Qian\inst{1} \and
Heng Zhou\inst{1,3}\textsuperscript{*} \and
Jiahao Zhang\inst{1} \and
Yajie Zhang\inst{1} \and
Zhihang Li\inst{1}\textsuperscript{\dag} \and
Mengyu Zhou\inst{1} \and
Erchao Zhao\inst{1} \and
Xiaoxi Jiang\inst{1} \and
Guanjun Jiang\inst{1}
}

\authorrunning{Y.~Liu et al.}

\institute{
Qwen Business Unit, Alibaba Group, China
\and
Huazhong University of Science and Technology, China
\and
Zhejiang University, China
}

\maketitle

\blfootnote{\textsuperscript{*} Work done during an internship at Alibaba.}
\blfootnote{\textsuperscript{\dag} Corresponding author.}

\begin{abstract}

Interleaved text-and-image generation represents a significant frontier for Multimodal Large Language Models (MLLMs), offering a more intuitive way to convey complex information. Current paradigms rely on either image generation or retrieval augmentation, yet they typically treat the two as mutually exclusive paths, failing to unify factuality with creativity. We argue that the next milestone in this field is \textit{Agentic Tool Planning}, where the model serves as a central controller that autonomously determines when, where, and which tools to invoke to produce interleaved responses for visual-critical queries. To systematically evaluate this paradigm, we introduce \name, a novel benchmark comprising 7,702 QA pairs (including 1,592 VQA pairs) across eight categories and 25 visual-critical intents, featuring human-verified queries and ground truths. Furthermore, to evaluate agentic planning independent of end-to-end execution and changing tool backends, we propose a Multi-Agent MLLM-as-a-Judge (MAM) system. MAM evaluates tool-call precision, identifies missed opportunities for tool use, and assesses overall response quality without requiring ground-truth references. Our extensive experiments on 10 state-of-the-art MLLMs reveal that models struggle with coherent interleaved planning and exhibit significant variations in tool-use behavior, highlighting substantial room for improvement and providing actionable guidance for advancing interleaved generation.

\end{abstract}
\section{Introduction}
\begin{figure}[t]
    \centering
    \includegraphics[width=\linewidth]{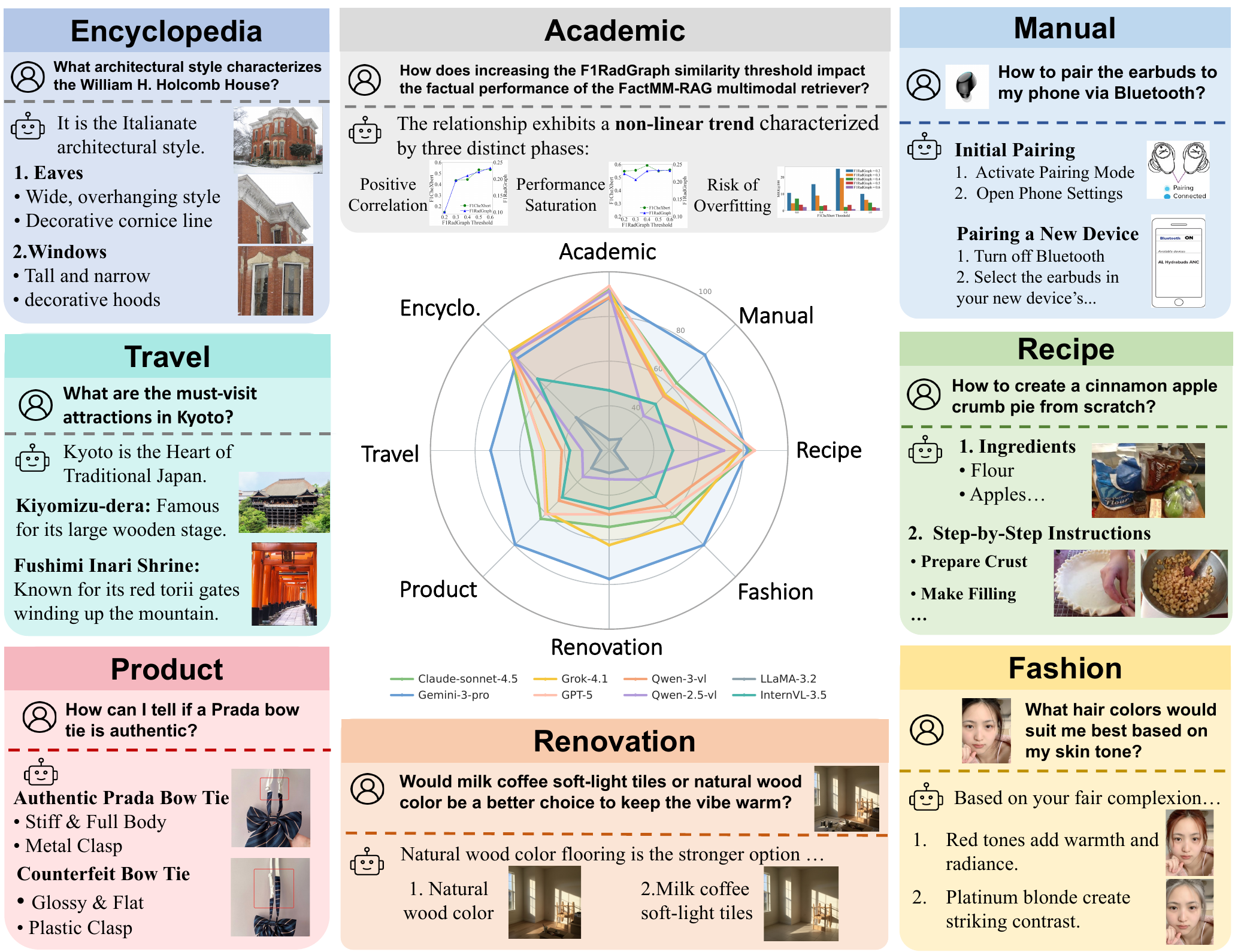}
    \caption{Examples of eight task categories and corresponding model performance. Since our benchmark targets agentic tool-planning ability, evaluation outputs and ground truths contain tool-planning tags instead of rendered images (See Figure~\ref{fig:overview}). For better interpretability, we present results obtained after end-to-end tool execution.}
    \label{fig:examples}
\end{figure}

Interleaved generation, which aims to jointly produce coherent sequences of text and images, represents a burgeoning frontier for multimodal large language models (MLLMs)~\cite{an2023openleaf,liu2024holistic,zhou2025opening,guo2025llm,xia2024mmie}. Unlike text-only responses, interleaved content provides a more intuitive and efficient way to convey information~\cite{mayer2002multimedia,taneja2025towards,yu2025mramg,zhang2025rag}. Figure~\ref{fig:examples} shows representative examples of this task. For instance, a researcher can interpret experimental results more effectively through figures than dense text (top panel); a user seeking styling advice benefits from previewing a modified hairstyle (bottom-right panel); and culinary instructions become more actionable with step-by-step visual aids (right panel). By grounding abstract explanations in concrete visual context, interleaved generation can minimize users' cognitive load and bridge the gap between digital assistance and real-world application~\cite{chen2024interleaved}.

Research on interleaved generation has largely converged to two separate paradigms.
The first paradigm focuses on \emph{image generation} via either unified models~\cite{team2024chameleon,chern2024anole,sun2023emu} or diffusion-based pipelines~\cite{podell2023sdxl,zhou2025opening,an2023openleaf}, but it often lacks factual grounding and struggles with complex diagrams~\cite{zhang2025rag}.
The second paradigm adopts \emph{retrieval augmentation} by retrieving and citing images from external corpora~\cite{zhang2025rag,yu2025mramg}, but it is inherently limited in its ability to modify visuals and support query-specific creative generation. 
However, neither paradigm captures a common real-world setting where the model must both reference context images and generate query-specific visuals within the same interleaved response, leaving this practical requirement largely unmet.

We argue that the next milestone for interleaved generation is \emph{Agentic Tool Planning}, which dissolves the boundary between reference and generation. In this paradigm, the model acts as a central controller, autonomously deciding \emph{when}, \emph{where}, and \emph{which} tools to invoke, dynamically orchestrating capabilities such as citing provided images, performing targeted edits, synthesizing new content, and acquiring real-world assets via web search to produce interleaved responses.

\begin{table}[t]
\centering
\caption{Comparison of \name with existing benchmarks. \name is the first to integrate hybrid image sourcing (Reference \& Generation) and dual query types (QA \& VQA) with expert-level annotations for both queries and ground truths.
}
\label{tab:dataset_comparison}
\resizebox{0.8\columnwidth}{!}{
\begin{tabular}{@{}lccccccc@{}}
\toprule
\multirow{2}{*}{\textbf{Dataset name}} 
& \multirow{2}{*}{\textbf{\#Sample}} 
& \multicolumn{2}{c}{\textbf{Query type}} 
& \multicolumn{2}{c}{\textbf{Response IMG source}} 
& \multirow{2}{*}{\textbf{\shortstack{Annotated\\Query}}} 
& \multirow{2}{*}{\textbf{\shortstack{Annotated\\GT}}} \\ 

\cmidrule(lr){3-4} 
\cmidrule(lr){5-6}

&  & QA & VQA & Reference & Generation &  &  \\ 
\midrule

OpenLEAF~\cite{an2023openleaf} & 30& \cmark & \xmark & \xmark & \cmark & \xmark & \xmark \\
InterleavedBench~\cite{liu2024holistic} &  815& \cmark & \cmark & \xmark & \cmark & \xmark & \xmark \\
OpenING~\cite{zhou2025opening} & 5,400& \cmark & \cmark & \xmark & \cmark & \xmark & \xmark \\
RAG-IGBench~\cite{zhang2025rag} & 6,057& \cmark & \xmark & \cmark & \xmark & \cmark & \cmark \\
MRAMG-Bench~\cite{yu2025mramg} & 4,800& \cmark & \xmark & \cmark & \xmark & \cmark & \cmark \\
LLM-I~\cite{guo2025llm} & 30& \cmark & \cmark & \xmark & \cmark & \xmark & \xmark \\ 
\midrule
\textbf{Ours (\name)} & \textbf{7,702}& \cmark & \cmark & \cmark & \cmark & \cmark & \cmark \\ 

\bottomrule
\end{tabular}
}
\end{table}

Studying this paradigm requires a dedicated benchmark that jointly models both reference and generation. In contrast, existing benchmarks typically evaluate only one side in isolation.
To bridge this gap, we present \name (\textbf{A}gentic \textbf{T}ool \textbf{P}lanning Bench), a novel benchmark for evaluating MLLMs on user-centric interleaved tool-planning tasks. As summarized in Table~\ref{tab:dataset_comparison}, \name is the first to jointly support hybrid image sourcing (reference \& generation) and dual query types (QA \& VQA), with expert-level annotations for both queries and ground truths. We ensure benchmark reliability through a systematic data collection and validation pipeline, along with a carefully designed evaluation strategy. We define a taxonomy of visual-critical categories and intents to guide query synthesis, with expert annotators verifying clarity and visual-criticality. Ground truth is constructed through a three-stage procedure with multi-perspective human evaluation. The resulting \name comprises 7,702 QA pairs, including 1,592 VQA pairs, across eight categories and 25 visual-critical intents. To evaluate agentic tool-planning ability, we further propose a Multi-Agent MLLM-as-a-Judge (MAM) system that assesses tool-call precision, missing tool-use opportunities, and overall response quality without requiring ground truth or end-to-end execution. The reliability of MAM is further validated through a human agreement study.

\noindent\textbf{Key Findings.}~Our comprehensive evaluation across 10 state-of-the-art reveals three key insights: (1) Existing MLLMs struggle to generate coherent interleaved tool plan, particularly for Travel and Renovation; (2) Gemini 3 Pro achieves the leading performance under our task setting; (3) Models exhibit distinct behavioral tendencies in terms of tool-call frequency and preference.

Our contributions are as follows: 
(1) We propose a novel paradigm, namely \emph{Agentic Tool Planning}, for interleaved generation, and introduce \name, a benchmark enabling the systematic study of tool planning capabilities under real-world interleaved settings for MLLMs.
(2) We propose a MAM system to assess tool-call precision and identify missed tool-use opportunities, without requiring ground truth or end-to-end execution. 
(3) We conduct extensive experiments on state-of-the-art MLLMs, revealing their tool-planning capabilities and potential biases, and providing actionable guidance for future research.
\section{Related Work}
\noindent\textbf{Methods for Interleaved Generation.}~
Interleaved generation produces coherent sequences that interleave textual reasoning with visual outputs. Early work explored unified autoregressive models that align text and image tokens end-to-end (e.g., Chameleon~\cite{team2024chameleon}, Show-o~\cite{xie2024show}, LLaVA-NeXT-Interleave~\cite{li2024llava}, Orthus~\cite{kou2024orthus}, Anole~\cite{chern2024anole}). In contrast, pipeline-based systems~\cite{wu2023visual,liu2024holistic,zhou2025opening,an2023openleaf,guo2025llm} decouple language modeling from visual synthesis, allowing LLMs to invoke specialized vision modules, while retrieval-augmented methods improve grounding by retrieving and citing images from external corpora~\cite{zhang2025rag,yu2025mramg}. Nonetheless, most prior work favors either generation or retrieval, rather than supporting both in one interleaved response.

\noindent\textbf{Evaluation for Interleaved Generation.}~Recent benchmarks have increasingly systematized interleaved-generation evaluation. General-purpose datasets (e.g., OpenLEAF~\cite{an2023openleaf}, MMIE~\cite{xia2024mmie}, InterleavedBench~\cite{liu2024holistic}, OpenING~\cite{zhou2025opening}, InterSyn~\cite{feng2025high}) typically use MLLM-as-a-judge~\cite{chen2024mllm} to assess image quality and cross-modal coherence. Retrieval-oriented benchmarks (RAG-IGBench~\cite{zhang2025rag}, MRAMG-Bench~\cite{yu2025mramg}) emphasize factual grounding via retrieval, measuring recall, precision, and semantic alignment. Specialized frameworks further extend MLLM-based evaluation to structured/interactive settings, including LLM-I~\cite{guo2025llm}, ISG-Bench~\cite{chen2024interleaved}, and WEAVE~\cite{chow2025weave}. Overall, existing benchmarks largely evaluate generation or retrieval in isolation with end-to-end metrics, prioritizing output fidelity and alignment over open-ended tool planning.

\noindent\textbf{Tool-Augmented MLLMs.}~
Recent advancements in tool-augmented MLLMs integrate external tools to enhance visual reasoning and action capabilities. Frameworks like ViperGPT~\cite{suris2023vipergpt} and MM-ReAct~\cite{yang2023mm} employ prompting pipelines for code and API execution in multimodal tasks, while AssistGPT~\cite{gao2023assistgpt}, LLaVA-Plus~\cite{liu2024llava}, and CLOVA~\cite{gao2024clova} utilize cyclic pipelines involving planning, execution, and refinement feedback. MLLM-Tool~\cite{wang2025mllm} adopts a learning-based approach with multimodal encoders for tool selection and agentic execution. Although tool-augmented MLLMs have inspired new paradigms for interleaved generation, such as LLM-I~\cite{guo2025llm}, the lack of valid datasets and benchmarks still hinders effective evaluation of their tool planning abilities.
\section{Task Formulation}

Our goal is to plan an interleaved multimodal response $R$ that tightly couples text with relevant images, given a visual-critical query $q$ and a document set $\mathcal{D}=\{d_1,d_2,\dots,d_n\}$. The set $\mathcal{D}$ is \emph{source-agnostic}: it may be retrieved from an external knowledge base, or provided directly by the user as context. Each document $d_i\in\mathcal{D}$ is represented as a tuple $(T_i,\mathcal{I}_i)$, where $T_i$ denotes the textual content and $\mathcal{I}_i$ is the set of images associated with the document. To generate $R$, an MLLM consumes a concatenated input comprising an interleaved generation prompt $p$, the query $q$, and the document set $\mathcal{D}$. The model then produces an ordered sequence:
\begin{equation}
    R=\{s_1,s_2,\dots,s_m\},\quad s_j\in\mathcal{S}_{\text{text}}\cup\mathcal{S}_{\text{tool}},
\end{equation}
where $\mathcal{S}_{\text{text}}$ denotes natural-language tokens and $\mathcal{S}_{\text{tool}}$ denotes tool-calling instructions for image integration.

\noindent\textbf{Visual-Critical Queries.}~
We define \emph{visual-critical queries} as prompts for which an interleaved image--text response offers substantially higher utility and information density than a text-only response. Such queries typically involve: 

\begin{itemize}
  \item \emph{Visual information augmentation}: images convey essential details that are difficult to capture precisely in language.
  \item \emph{Cognitive acceleration}: visuals exploit parallel perception to speed up comprehension, especially in spatially intensive tasks and hands-on procedures.
  \item \emph{Structural illustration}: diagrams make structure and relationships explicit.
\end{itemize}

\noindent\textbf{Toolkit.}~
To support principled visual integration within the interleaved response, we define a unified tool-calling space $\mathcal{S}_{\text{tool}}$. Each tool invocation follows a structured schema:
\texttt{<tool>\{"tool\_name": \dots, "description": \dots, "params": \dots\}</tool>}.
The toolkit comprises five specialized modules that cover complementary needs for visual acquisition and manipulation:

\begin{itemize}
    \item \emph{Reference}: anchors the response to specific visual evidence in $\mathcal{D}$ by specifying an \texttt{img\_index}, enabling faithful citation of in-context images.

    \item \emph{Diffusion}: synthesizes novel images for conceptual illustrations or artistic renderings absent from $\mathcal{D}$. It takes a semantically detailed \texttt{prompt} describing the intended content.

    \item \emph{Search}: retrieves real-world visuals via external search engines such as Google image search. It accepts a targeted search \texttt{query} to ground the response in factual, up-to-date imagery.

    \item \emph{Code}: generates programmatic, data-driven visualizations such as charts and mathematical plots. It requires a diagram \texttt{type} and a specification of the \texttt{data}, ensuring precision and interpretability.

    \item \emph{Edit}: modifies a referenced image from $\mathcal{D}$ for localized refinement. Given an \texttt{img\_index} and an edit \texttt{prompt}, it can add labels, highlight regions, or annotate salient visual features.
\end{itemize}

\section{\name}
\begin{figure}[t]
    \centering
    \begin{subfigure}[t]{\linewidth}
        \centering
        \includegraphics[width=\linewidth]{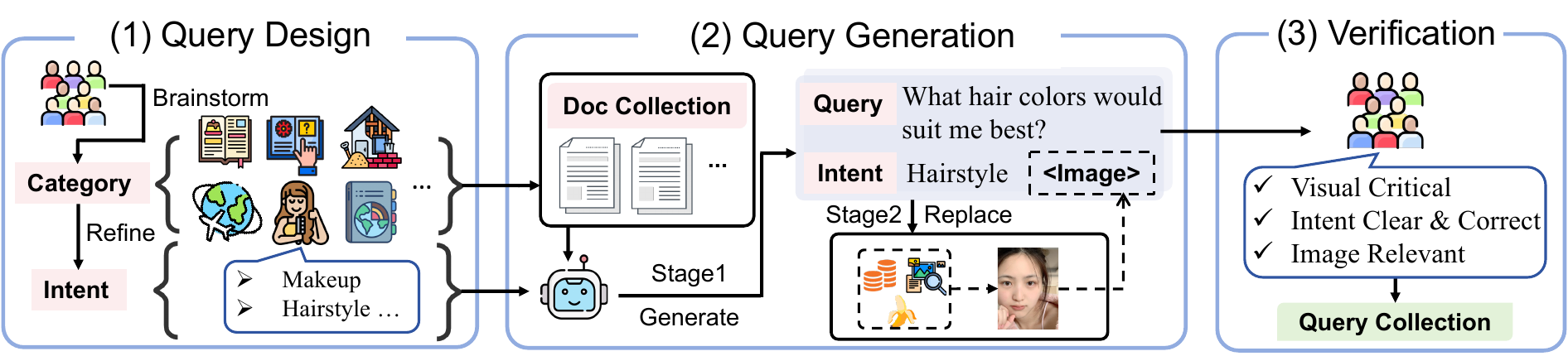}
        \caption{Query collection process of \name.}
        \label{fig:query_construction}
    \end{subfigure}
    
    \vspace{0.5em}
    
    \begin{subfigure}[t]{\linewidth}
        \centering
        \includegraphics[width=\linewidth]{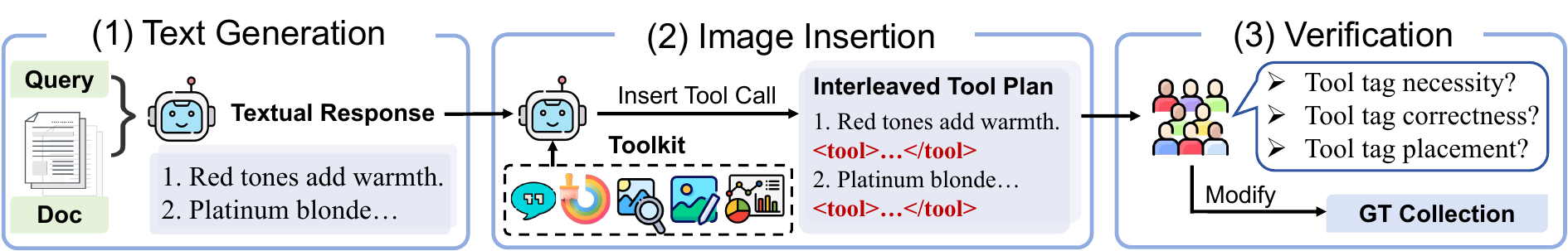}
        \caption{Ground truth collection process of \name.}
        \label{fig:gt_construction}
    \end{subfigure}
    
    \vspace{0.8em}
    
    \begin{subfigure}[t]{\linewidth}
        \centering
        \includegraphics[width=\linewidth]{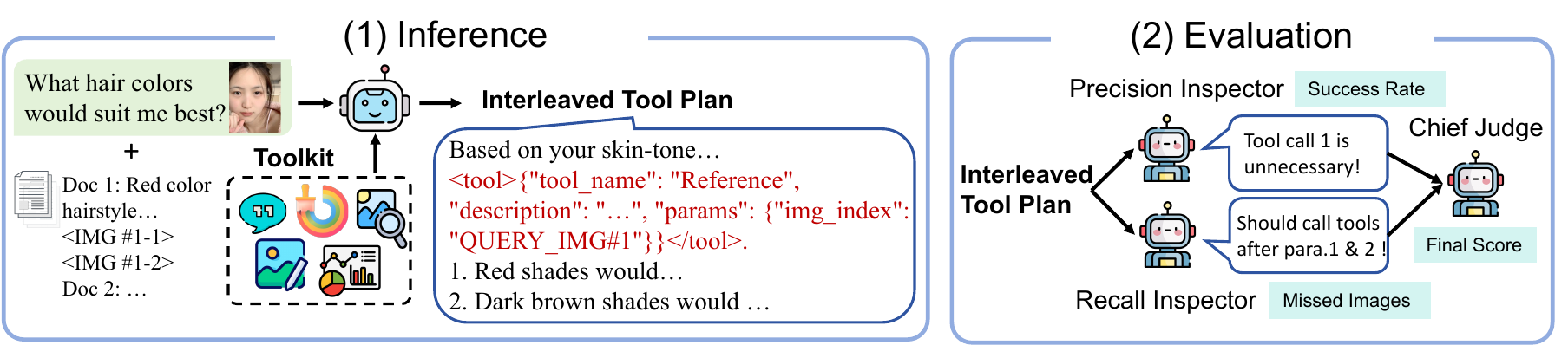}
        \caption{Inference and evaluation process of \name.}
        \label{fig:inference_eval}
    \end{subfigure}
    
    \caption{Overview of \name dataset construction and evaluation pipelines.}
    \label{fig:overview}
\end{figure}

\begin{table}[t] 
\centering 
\caption{Statistics of \name. Note that "\#TC per GT" stands for the number of tool calls per ground truth answer.} 
\label{tab:dataset_stats} 
\resizebox{0.9\textwidth}{!}{ 
\begin{tabular}{l l R{1cm} R{1.45cm} R{1.45cm} R{1.45cm}} \toprule 
\textbf{Category} & \textbf{Intent} & \textbf{\#Query} & \makecell[r]{\textbf{\#Doc} \\ \textbf{per Q}} & \makecell[r]{\textbf{\#Img} \\ \textbf{per Q}} & \makecell[r]{\textbf{\#TC} \\ \textbf{per GT}} \\ \midrule 
\textbf{Academic} & Framework, Component, Comparison, Dataset, Result & 196 & 1.00 & 3.34 & 1.96 \\
\textbf{Manual} & Operation Guide, Function, Component & 1,025 & 2.53 & 10.90 & 4.38 \\ 
\textbf{Recipe} & Guideline, Ingredient & 1,960 & 1.23 & 6.25 & 3.06 \\ 
\textbf{Fashion} & Hairstyle, Makeup, Outfit, Photography & 1,051 & 2.95 & 11.81 & 4.63 \\ 
\textbf{Renovation} & Element, Style & 440 & 2.92 & 11.98 & 4.79 \\ 
\textbf{Product} & ProductIntro, Comparison, Authenticity & 1,330 & 2.91 & 10.40 & 4.58 \\ 
\textbf{Travel} & Planning, Navigation & 734 & 2.91 & 16.00 & 8.00 \\ 
\textbf{Encyclopedia} & Animals, Geography, Biography, Architecture & 966 & 1.01 & 1.05 & 3.65 \\ \midrule 
\textbf{Overall} & \textbf{--} & \textbf{7,702} & \textbf{2.15} & \textbf{8.87} & \textbf{4.32} \\ \bottomrule 
\end{tabular}} 
\end{table}
This section provides a comprehensive overview of \name. \name dataset consists of 7,702 QA pairs (including 1,592 VQA pairs) spanning eight categories and 25 visual-critical intents. The detailed statistics are provided in Table~\ref{tab:dataset_stats}. We detail the construction methodologies for both the query and ground truth collections as illustrated in Figure~\ref{fig:query_construction} and~\ref{fig:gt_construction}. Subsequently, we introduce MAM, the evaluation system specifically designed for this task, as shown in the right panel of Figure~\ref{fig:inference_eval}.

\subsection{Query Collection}
\noindent\textbf{Design.}~We identify eight \textbf{high-visual-demand query categories} in which users are most likely to expect image-rich responses: \textit{Academic}, \textit{Manual}, \textit{Recipe}, \textit{Fashion}, \textit{Renovation}, \textit{Product}, \textit{Travel}, and \textit{Encyclopedia}. For each category, we collect visually grounded documents from existing multimodal benchmarks, including MRAMG-Bench, RAG-IGBench, and OVEN~\cite{hu2023open}, which aggregate content from diverse sources such as Wikipedia, Xiaohongshu, and arXiv. We further define fine-grained \textbf{visual-critical intents} that intrinsically require visual support, resulting in \textbf{25 distinct intents} spanning the eight categories.

\noindent\textbf{Generation.}~We employ Gemini 2.5 Pro~\cite{comanici2025gemini} to generate queries via a two-stage pipeline. In the first stage, given a source document and its predefined intent taxonomy, the model identifies the most relevant intent(s) and synthesizes \emph{text-only} queries conditioned on them. The prompts encourage a \emph{natural, conversational tone} that resembles everyday user phrasing rather than explicit, tool-oriented requests. In the second stage, building on these text queries, we select intents that can be visually grounded and convert them into VQA-style queries by replacing key textual evidence with images. For such visually grounded queries, e.g., fashion advice based on user portraits, interior renovation suggestions based on photos, or encyclopedic questions about animals and landmarks, we curate a high-quality visual corpus from three sources, including web search, generative synthesis using \textit{nano-banana}~\cite{nanobanana}, and public benchmarks such as OVEN.

\noindent\textbf{Verification.}~
We employed a team of ten professional annotators to verify the quality of the generated queries. During this process, annotators removed queries with (i) ambiguous or unnatural queries, (ii) VQA pairs where the question is not grounded in the associated image, or (iii) queries that do not meet our definition of visual-critical queries. After filtering, approximately 95\% of queries were retained. The queries are also relabeled with their corresponding categories and intent labels to ensure correctness, enabling reliable fine-grained analysis.

\subsection{Ground Truth Collection}
We used a three-stage process for ground truth generation. 

\noindent\textbf{Textual Response Generation.}~
We prompt an MLLM to generate a high-quality text-only answer grounded in the provided documents and images. The prompt enforces factual consistency with document text, accurate visual interpretation of the query and images, and coherent reasoning with complete query coverage. It also imposes task-specific formatting requirements, such as step-by-step instructions or a structured introduction with multi-aspect comparisons, and mandates an answer-first, concise style without filler.

\noindent\textbf{Image Insertion.}~
To address visual gaps, we instruct an MLLM to proactively invoke specialized visual tools and place the generated outputs immediately after the relevant paragraph. Tool usage follows a clear \textbf{capability boundary}. \emph{Reference} supports multi-step procedures, complex diagrams, and context-dependent examples. \emph{Search} retrieves real-world entity images such as landmarks, artworks, notable figures, events, and maps. \emph{Diffusion} generates creative concepts, design drafts, abstract visuals, and fashion demonstrations when no suitable reference exists. \emph{Edit} modifies provided images, including highlighting, annotation, renovation mock-ups, and style previews. \emph{Code} produces data-driven visualizations and blueprint-style diagrams. We further enforce strict integration rules: no figure-introducing phrases, single use per reference image, no mid-sentence insertion, and a high-value visualization policy to ensure clarity and coherence.

\noindent\textbf{Fine-grained Annotation and Refinement.}
To ensure dataset fidelity and trustworthiness, we asked 15 annotators to review every tool call. A tool call is retained only if its expected image materially improves understanding, uses correct parameters, and is placed immediately after the relevant text to preserve narrative coherence. Annotators remove redundant or semantically mismatched calls, fix issues such as malformed syntax and inappropriate tool choices, and relocate misplaced tags to paragraph boundaries to avoid disrupting readability. We discard the entire sample when errors are systemic, such as severe markdown structure breakage or fundamental factual inaccuracies in the text.

\subsection{MAM: Multi-Agent MLLM-as-a-judge}

Evaluating model performance in our paradigm introduces challenges that conventional metrics cannot adequately address. First, the open-ended nature of the tasks means that ground truths function as high-quality references rather than definitive standards for every tool invocation. Second, we emphasize evaluating tool planning rather than end-to-end execution, thereby isolating the model’s intrinsic planning ability from external API constraints. Third, evaluation must be multi-dimensional, assessing both the precision of tool calls, including their necessity, correctness, and placement, and the recall of missed visual elements when such support is required. To address these challenges, we propose a Multi-Agent MLLM-as-a-Judge system that evaluates tool-call precision, missed tool-use opportunities, and overall response quality without relying on ground truth or full execution. The framework comprises three specialized agents: the Precision Inspector, the Recall Inspector, and the Chief Judge.

\noindent\textbf{The Precision Inspector.}~
The Precision Inspector evaluates the precision and execution quality of each tool invocation in the model response using a two-stage procedure. It first verifies visual-critical necessity, requiring that the image provides clear added value over text, and checks tool boundary compliance to ensure the selected tool matches its technical capabilities. For invocations that pass these prerequisites, it then assesses semantic placement, structural coherence with surrounding content, parameter accuracy, and output format correctness, and assigns a score on a 0–2 scale where 0 indicates failure, 1 indicates partial satisfaction, and 2 indicates full satisfaction.

\noindent\textbf{The Recall Inspector.}~
The Recall Inspector identifies missed tool-use opportunities where the model responds with text only but should have invoked a tool. It flags cases where the text explicitly refers to an absent visual, where long textual descriptions of complex procedures, reference images, or real-world entities lack appropriate \emph{reference} or \emph{search} grounding, and where the model omits \emph{diffusion}, \emph{edit}, or \emph{code} despite an implicit need for creative generation, image modification, or plotting.

\noindent\textbf{The Chief Judge.}~
The Chief Judge synthesize the overall interleaved pacing by integrating the Precision and Recall Reports into a holistic, quantitative score ranging from 0 to 100. This scoring mechanism is strictly anchored across five performance tiers: 
(i) \textit{Excellent (80--100)} represents near-perfect execution, where all tools pass necessity and boundary checks with precise syntax and zero missed visual opportunities; 
(ii) \textit{Good but Flawed (60--80)} denotes a genuinely helpful response characterized by minor precision issues or at most one minor visual omission; 
(iii) \textit{Mediocre (40--60)} indicates noticeable discrepancies, including multiple parameter errors, forced structural splits, or 1--2 missed opportunities identified by the Recall Inspector; 
(iv) \textit{Poor (20--40)} signifies significant failures such as severe visual redundancy, tool boundary violations, or 2--3 instances where explicit visual contexts were improperly relegated to plain text; and 
(v) \textit{Fatal (0--20)} reflects a complete breakdown of execution, marked by severe formatting disruptions, or the total omission of the user's core visual intent. 
\section{Experiment}
\subsection{Setup}

\noindent\textbf{Inference Settings.}~We evaluate 10 MLLMs, namely Claude Sonnet 4.5~\cite{claude4.5}, Claude Sonnet 4~\cite{claude4}, Gemini 3 Pro~\cite{gemini3}, Grok-4.1 Fast Reasoning~\cite{grok4.1}, GPT-5~\cite{gpt5}, GPT-4o~\cite{hurst2024gpt}, Qwen3-VL-Plus~\cite{bai2025qwen3}, Qwen2.5-VL-72B~\cite{bai2025qwen25vltechnicalreport}, LLaMA-3.2-11B~\cite{llama}, InternVL3.5-14B~\cite{wang2025internvl3}. The prompt for interleaved generation is provided in the supplementary materials, which also use the same tool boundary as in the prompt for ground-truth generation. By default, we use a zero-shot strategy to evaluate the model performance. In addition, we conduct a few-shot (3-shot) experiment on Claude Sonnet 4.5, GPT-4o, Qwen2.5-VL-72B, LLaMA-3.2-11B.

\noindent\textbf{MAM Settings.}~By default, we use Gemini 2.5 Pro as the agents in the MAM framework. We extend the agents to Claude Sonnet 4.5 and GPT-5 in our ablation study. The prompt for agents is provided in the supplementary materials.

\subsection{Evaluation Metrics}
\noindent\textbf{Final Score (FS).}~The Final Score refers to the score assigned by the Chief Judge in our MAM system. It reflects the overall quality of the response.

\noindent\textbf{Success Rate (SR).}~The Success Rate measures tool-call execution accuracy based on the Precision Inspector. A tool call is considered successful only if it is necessary, semantically appropriate, well-placed, and uses the correct tool with the correct parameters in the proper format.

\noindent\textbf{Missed Images (MI).}~The Missed Images count is reported by the Recall Inspector, who reviews the response and identifies missed opportunities for tool calls based on omission criteria. Lower missed image counts mean the models are more capable of filling visual-critical gaps.

\noindent\textbf{Tool Adoption Rate.}~Thw Tool Adoption Rate measures the percentage of queries in which the model invokes a specific tool at least once. It reflects the model’s overall preference for incorporating a specific tool into responses.

\noindent\textbf{Precision, Recall, and F1-score.}~We also evaluate tool invocation by comparing model-selected tool sets against our curated ground truth. Precision and Recall measure the accuracy and completeness of tool selection, respectively, while the F1-score provides a balanced assessment of overall performance.

\subsection{Results}

\begin{table}[t]
\centering
\caption{Final Score of models across eight categories (Renovation and Encyclopedia is abbreviated as Renova. and Encyclo. respectively).}
\label{final score}
\resizebox{0.9\textwidth}{!}{ %
\begin{tabular}{@{}lccccccccc@{}}
\toprule
 \textbf{MLLM} & \textbf{Academic} & \textbf{Manual} & \textbf{Recipe} & \textbf{Fashion} & \textbf{Renova.} & \textbf{Product} & \textbf{Travel} & \textbf{Encyclo.} & \textbf{Avg} \\ \midrule
\textbf{Claude Sonnet 4.5} & \underline{91.93} & 62.74 & \underline{83.86} & 61.79 & 54.22 & 63.25 & 54.56 & 82.39 & \underline{69.34} \\
\textbf{Claude Sonnet 4} & 88.12 & \underline{63.37} & 82.85 & 62.21 & 54.33 & \underline{63.87} & \underline{54.77} & \textbf{83.69} & 69.15 \\
\textbf{Gemini 3 Pro} & 88.32 & \textbf{80.58} & 81.96 & \textbf{79.83} & \textbf{77.53} & \textbf{79.51} & \textbf{73.07} & 78.20 & \textbf{79.88} \\
\textbf{Grok-4.1} & 91.43 & 55.39 & 78.83 & \underline{66.05} & \underline{62.37} & 59.20 & 49.18 & \underline{82.96} & 68.18 \\
\textbf{GPT-5} & \textbf{93.59} & 61.00 & \textbf{85.13} & 57.91 & 48.63 & 60.31 & 49.63 & 81.27 & 67.18 \\
\textbf{GPT-4o} & 86.27 & 56.38 & 79.86 & 48.76 & 43.19 & 53.34 & 43.94 & 73.59 & 60.67 \\
\textbf{Qwen3-VL-Plus} & 88.51 & 54.42 & 78.87 & 54.99 & 48.67 & 51.73 & 41.22 & 81.51 & 62.49 \\
\textbf{Qwen2.5-VL-72B} & 91.36 & 41.90 & 71.32 & 38.49 & 32.90 & 36.61 & 32.06 & 81.11 & 53.22 \\ 
\textbf{InternVL3.5-14B} & 46.90 & 49.34 & 48.52 & 49.24 & 46.06 & 49.68 & 37.58 & 65.37 & 49.09 \\ 
\textbf{LLaMA-3.2-11B} & 24.80 & 27.34 & 22.34 & 31.48 & 30.08 & 31.15 & 23.64 & 40.94 & 28.97 \\ \bottomrule
\end{tabular}
} %
\end{table}

\begin{table}[t]
\centering
\caption{Success Rate of models across eight categories.}
\label{success rate}
\resizebox{0.9\textwidth}{!}{ %
\begin{tabular}{@{}lccccccccc@{}}
\toprule
 \textbf{MLLM} & \textbf{Academic} & \textbf{Manual} & \textbf{Recipe} & \textbf{Fashion} & \textbf{Renova.} & \textbf{Product} & \textbf{Travel} & \textbf{Encyclo.} & \textbf{Avg} \\ \midrule
\textbf{Claude Sonnet 4.5} & \underline{92.09} & \underline{66.95} & \underline{81.95} & 71.55 & 65.10 & 71.44 & \underline{72.98} & \textbf{84.06} & \underline{75.77} \\
\textbf{Claude Sonnet 4}   & 87.31 & 66.60 & 80.49 & \underline{71.80} & 62.58 & \underline{71.46} & 70.77 & 79.85 & 73.86 \\
\textbf{Gemini 3 Pro}      & 78.89 & \textbf{82.97} & 74.11 & \textbf{85.92} & \textbf{86.43} & \textbf{86.44} & \textbf{87.86} & 71.55 & \textbf{81.77} \\
\textbf{Grok-4.1}          & 90.82 & 57.92 & 74.53 & 69.77 & \underline{71.41} & 60.92 & 62.54 & \underline{81.79} & 71.21 \\
\textbf{GPT-5}             & \textbf{94.54} & 60.80 & \textbf{84.41} & 64.92 & 47.62 & 66.43 & 63.08 & 72.01 & 69.23 \\
\textbf{GPT-4o}            & 85.83 & 55.24 & 79.13 & 54.07 & 38.34 & 63.56 & 55.96 & 60.89 & 61.63 \\
\textbf{Qwen3-VL-Plus}     & 85.63 & 55.81 & 78.01 & 62.65 & 48.39 & 56.60 & 50.78 & 73.57 & 63.93 \\
\textbf{Qwen2.5-VL-72B}    & 83.24 & 28.64 & 57.56 & 31.35 & 19.51 & 24.80 & 21.16 & 46.81 & 39.13 \\
\textbf{InternVL3.5-14B}  & 11.23 & 10.73 & 6.10 & 11.73 & 9.94 & 11.54 & 10.89 & 8.77 & 10.12 \\ 
\textbf{LLaMA-3.2-11B}     & 0.00 & 20.23 & 16.64 & 27.27 & 24.14 & 25.44 & 19.79 & 13.19 & 18.34 \\ \bottomrule
\end{tabular}
}
\end{table}

\begin{table}[t]
\centering
\caption{Missed Image of models across eight categories; lower is better.}
\label{missed images}
\resizebox{0.9\textwidth}{!}{ 
\begin{tabular}{@{}lccccccccc@{}}
\toprule
 \textbf{MLLM} & \textbf{Academic} & \textbf{Manual} & \textbf{Recipe} & \textbf{Fashion} & \textbf{Renova.} & \textbf{Product} & \textbf{Travel} & \textbf{Encyclo.} & \textbf{Avg} \\ \midrule
\textbf{Claude Sonnet 4.5} & 0.10 & 1.53 & 0.29 & 1.29 & 1.78 & 1.31 & 2.93 & 0.30 & 1.19 \\
\textbf{Claude Sonnet 4} & \underline{0.06} & \underline{1.18} & \textbf{0.23} & 1.09 & 1.55 & \underline{1.15} & \underline{2.38} & 0.25 & \underline{0.99} \\
\textbf{Gemini 3 Pro} & 0.15 & \textbf{0.44} & \underline{0.25} & \textbf{0.46} & \textbf{0.52} & \textbf{0.52} & \textbf{1.31} & \underline{0.24} & \textbf{0.49} \\
\textbf{Grok-4.1} & \textbf{0.05} & 1.55 & 0.44 & \underline{0.96} & \underline{1.22} & 1.27 & 2.75 & \textbf{0.22} & 1.06 \\
\textbf{GPT-5} & 0.14 & 1.32 & 0.31 & 1.30 & 1.67 & 1.29 & 2.60 & 0.26 & 1.11 \\
\textbf{GPT-4o} & 0.28 & 1.50 & 0.51 & 1.65 & 1.80 & 1.68 & 3.03 & 0.36 & 1.35 \\
\textbf{Qwen3-VL-Plus} & 0.17 & 1.75 & 0.55 & 1.70 & 1.98 & 1.83 & 3.25 & 0.31 & 1.44 \\
\textbf{Qwen2.5-VL-72B} & 0.15 & 2.11 & 0.70 & 2.00 & 2.21 & 2.38 & 3.65 & 0.27 & 1.68 \\
\textbf{InternVL3.5-14B} & 0.98 & 1.84 & 1.65 & 1.49 & 1.71 & 1.69 & 3.28 & 0.34 & 1.62 \\ 
\textbf{LLaMA-3.2-11B} & 1.37 & 2.32 & 2.03 & 2.07 & 2.26 & 2.26 & 3.82 & 1.11 & 2.16 \\ \bottomrule
\end{tabular}
} 
\end{table}

\noindent\textbf{Main Results}
We report results on tool-planning quality using three metrics: final score (Table~\ref{final score}), success rate (Table~\ref{success rate}), and missed images (Table~\ref{missed images}). 

\textbf{Gemini 3 Pro achieves leading performance across all metrics and most categories.}
Gemini 3 Pro attains the highest average final score of 79.88 and tool-use success rate of 81.77, while maintaining the lowest missed image count at 0.49. It shows clear advantages in tool-intensive domains, consistently reaching final scores around 77–81 with success rates above 82 in Manual, Fashion, Renovation, and Product. Even in Travel, it achieves the highest success rate of 87.86 while sustaining competitive overall performance.

\textbf{Tier-2 models achieve comparable final scores but exhibit distinct tool-use trade-offs.}
Claude Sonnet 4.5, Claude Sonnet 4, Grok-4.1, and GPT-5 obtain similar average final scores, yet differ noticeably in tool-use behavior. Claude Sonnet 4.5 achieves a higher success rate than Claude Sonnet 4, but also incurs more missed images, suggesting a more conservative tool-use strategy. GPT-5 performs strongly in language-intensive domains such as Academic, but degrades substantially in tool-heavy categories, with lower final scores and success rates in Renovation and Travel, accompanied by increased missed images.

\textbf{GPT-4o and Qwen3-VL-Plus form a mid-tier with weaker visual grounding coverage and execution robustness.}
GPT-4o and Qwen3-VL-Plus obtain similar average final scores, 60.67 and 62.49, but lag behind Tier-2 models in success rate and exhibit higher missed-image counts. The gap is most evident in tool-intensive categories such as Travel and Renovation, where both models show reduced success rates and substantially elevated missed images. Overall, their performance is constrained by insufficient visual gap detection and less robust tool execution when image use is required.

\textbf{Open-source models underperform overall, with heterogeneous failure patterns.}
Qwen2.5-VL-72B performs well in knowledge-intensive categories such as Academic and Encyclopedia, but drops sharply in tool-heavy domains, accompanied by a low average success rate and elevated missed images, indicating limited tool planning ability. InternVL3.5-14B shows fewer missed images than LLaMA-3.2-11B but an extremely low success rate, suggesting frequent invalid tool calls due to formatting issues. LLaMA-3.2-11B performs worst overall, with low final scores, low success rates, and high missed images, reflecting compounded weaknesses in both visual gap detection and valid tool execution.

\textbf{Across categories, Academic and Encyclopedia are the easiest, while Travel and Renovation are the hardest.}
Averaged across models, Academic and Encyclopedia achieve the highest final scores with the fewest missed images, reflecting limited need for complex tool coordination. In contrast, Travel is the weakest category overall, characterized by the lowest final scores across most models and substantially elevated missed-image counts, indicating failures in identifying tool-use opportunities. Renovation is also consistently challenging, but for a different reason: several models exhibit markedly reduced success rates despite only moderately increased missed images, suggesting that errors stem more from incorrect tool execution. Overall, Travel is primarily MI-dominated, whereas Renovation is more SR-dominated.

\begin{figure}[t]
    \centering
    \includegraphics[width=0.42\textwidth]{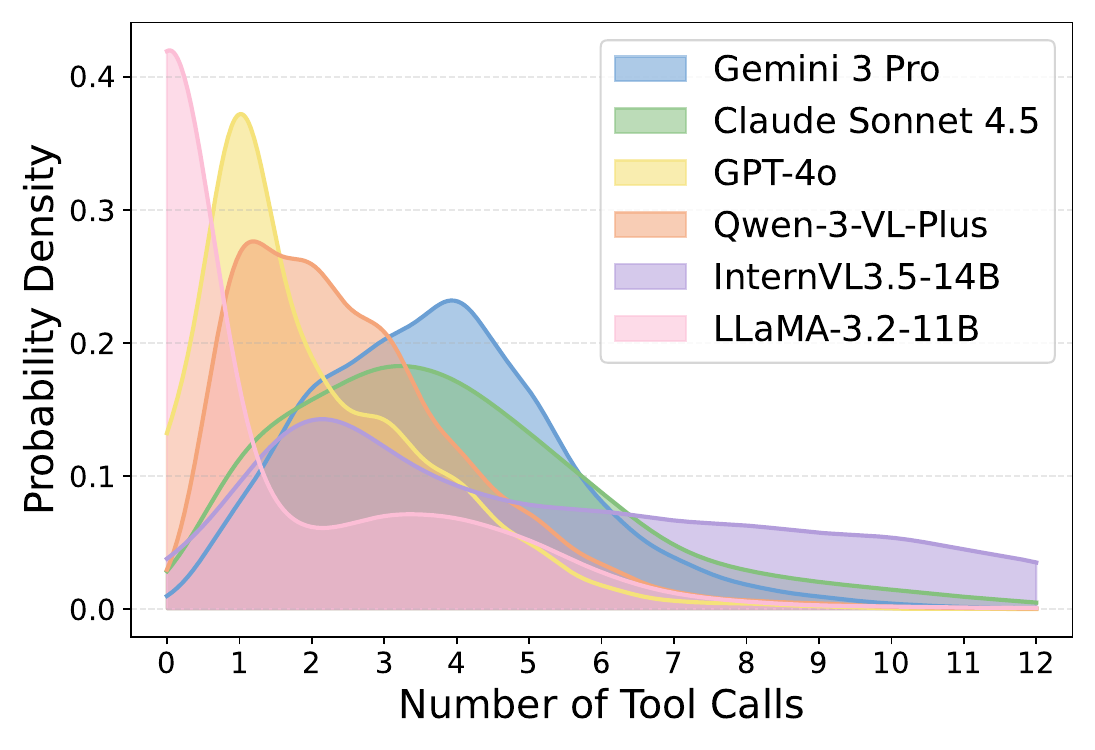}
    \caption{Tool call number distribution for representative models.}
    \label{fig:tool-call-num-dis-sample}
\end{figure}

\noindent\textbf{Tool Call Number Distribution.}~Figure~\ref{fig:tool-call-num-dis-sample} illustrates distinct tool-use strategies across models. 
Top-performing models cluster around three to five calls, indicating calibrated planning that avoids both under-invocation and redundancies. 
Mid-tier models skew toward fewer calls, often peaking around one to three, reflecting conservative triggering behavior that aligns with their higher missed-image counts in tool-heavy categories.
In contrast, smaller open-source models exhibit more extreme behaviors: LLaMA-3.2-11B concentrates mass at zero calls, suggesting systematic under-use and MI-dominated errors, while InternVL3.5-14B shows a long tail toward high call counts, consistent with SR-dominated failures due to difficulties in producing valid tool calls.

\begin{figure}[t] 
    \centering
    \includegraphics[width=0.8\textwidth]{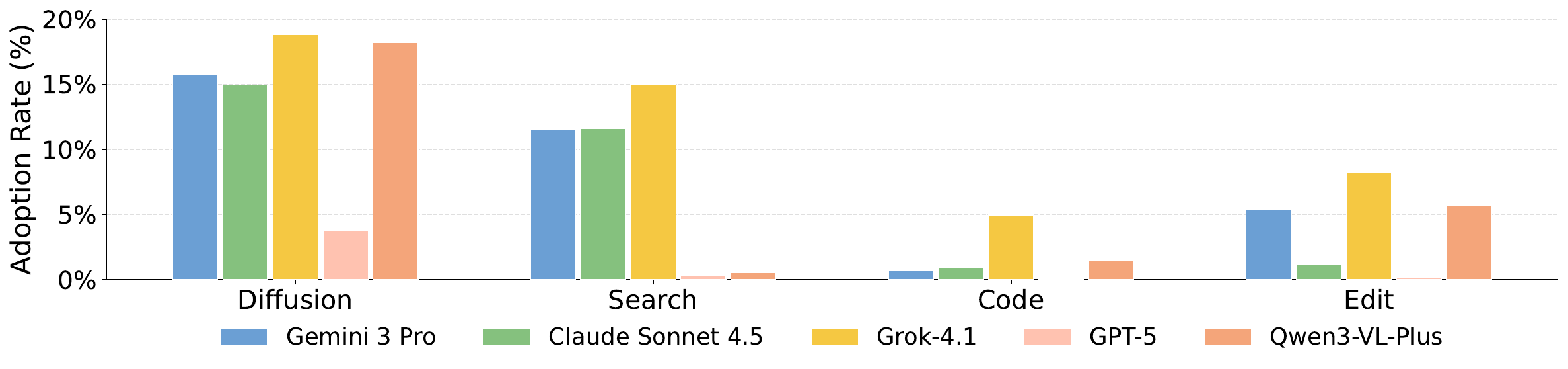} 
    \caption{Tool Adoption Rates per query for representative models. The reference tool is omitted due to its high dominance; all models utilize it in over 90\% of queries.}
    \label{fig:tool-adoption} 
\end{figure}

\begin{figure}[t] 
    \centering
    \includegraphics[width=0.8\textwidth]{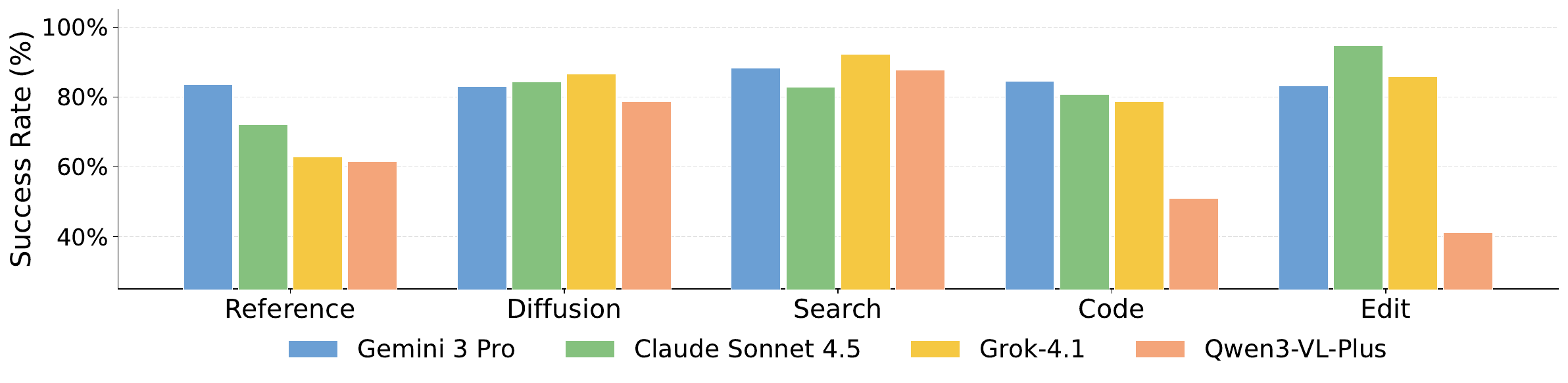} 
    \caption{Tool Success Rates per tool call for representative models on each tool.}
    \label{fig:tool-success} 
\end{figure}

\noindent\textbf{Tool Adoption Rate.}~Figure~\ref{fig:tool-adoption} reports per-query adoption rates of non-reference tools. \emph{Diffusion} is the most frequently invoked tool across models, followed by \emph{search}, while \emph{edit} is used moderately and \emph{code} is rarely employed. Beyond this overall trend, models display distinct strategies. Gemini 3 Pro shows a balanced distribution across \emph{diffusion}, \emph{search}, and \emph{edit}, aligning with its strong planning ability. Claude Sonnet 4.5 adopts a more conservative approach with lower auxiliary tool usage. Grok-4.1 is the most tool-active and diverse, including comparatively higher code usage, reflecting a more aggressive externalization strategy. GPT-5 relies minimally on non-reference tools, while Qwen3-VL-Plus primarily favors \emph{diffusion} and \emph{edit}, suggesting a generation-oriented interaction pattern.

\noindent\textbf{Tool Success Rate.}~
Figure~\ref{fig:tool-success} shows per-call success rates by tool and model. Overall, \emph{search} and \emph{diffusion} are most reliable, while \emph{code} and \emph{edit} show larger gaps. Gemini 3 Pro maintains consistently high SR across all tools, indicating strong tool-call generation. Claude Sonnet 4.5 performs similarly well and achieves the best results on \emph{edit}, reflecting robust editing-oriented interactions. Grok-4.1 excels on \emph{search} and remains competitive on \emph{diffusion} and \emph{edit}, but is weaker on \emph{reference}. In contrast, Qwen3-VL-Plus performs well on \emph{diffusion} and \emph{search} but drops significantly on \emph{code} and \emph{edit}, suggesting difficulty in generating correct tool calls in code- or edit-intensive scenarios.

\begin{table}[htbp]
    \centering
    \begin{minipage}[t]{0.48\textwidth}
        \centering
        \caption{Tool set precision, recall and F1-score evaluated by ground truth.}
        \label{tab:eval_gt}
        \resizebox{0.9\linewidth}{!}{
        \begin{tabular}{lccc}
            \toprule
            \textbf{Model} & \textbf{Precision} & \textbf{Recall} & \textbf{F1-score} \\
            \midrule
            \textbf{Claude Sonnet 4.5}   & \underline{89.64} & 78.15 & 80.49 \\
            \textbf{Claude Sonnet 4 }    & 88.05 & \underline{79.95} & \underline{80.71} \\
            \textbf{Gemini 3 Pro}        & 89.30 & 79.63 & \textbf{81.21} \\
            \textbf{Grok-4.1}            & 85.86 & \textbf{81.69} & 80.68 \\
            \textbf{GPT-5 }              & \textbf{91.42} & 69.81 & 76.37 \\
            \textbf{GPT-4o }             & 80.16 & 61.31 & 67.12 \\
            \textbf{Qwen3-VL-Plus}      & 85.93 & 73.31 & 75.58 \\
            \textbf{Qwen2.5-VL-72B}      & 88.87 & 72.28 & 76.53 \\
            \textbf{InternVL3.5-14B}     & 5.00  & 3.82  & 4.13  \\
            \textbf{LLaMA-3.2-11B}       & 12.42 & 13.58 & 11.98 \\
            \bottomrule
        \end{tabular}
        }
    \end{minipage}
    \hfill 
    \begin{minipage}[t]{0.45\textwidth}
        \centering
        \caption{Few-shot experiments on three models.}
        \label{fewshot}
        \resizebox{0.9\linewidth}{!}{
        \begin{tabular}{lccc}
            \toprule
            \textbf{Method}  & \textbf{FS}$\uparrow$ & \textbf{SR}$\uparrow$ & \textbf{MI}$\downarrow$\\ 
            \midrule
            \textbf{GPT-4o} & 60.35 & 63.05 & 1.39 \\
            \textbf{GPT-4o + 3 Shots} & \textbf{73.19} & \textbf{82.01} & \textbf{0.83} \\ 
            \midrule
            \textbf{Qwen2.5} & 53.88 & 39.06 & 1.64 \\
            \textbf{Qwen2.5+ 3 Shots} & \textbf{72.86} & \textbf{72.88} & \textbf{0.65} \\ 
            \midrule
            \textbf{LLaMA-3.2} & 29.60 & 25.35 & 2.10 \\
            \textbf{LLaMA-3.2 + 3 Shots} & \textbf{30.20} & \textbf{26.62} & \textbf{2.01} \\ 
            \bottomrule
        \end{tabular}
        }
    \end{minipage}
\end{table}

\noindent\textbf{Ground Truth Based Evaluation.}~To assess tool-invocation accuracy and completeness, we report Precision, Recall, and F1-score by comparing inferred and ground truth tool sets. This metric accounts for the fact that while exact execution traces may diverge, the core set of tools necessary to resolve a query should remain consistent. As shown in Table~\ref{tab:eval_gt}, strong models indicated by our MAM system (e.g., Gemini 3 Pro, Claude, Grok-4.1) achieve F1-scores above 80\%. We further compare the F1-scores ranking with our Final Score and find a high Spearman correlation ($\rho=0.879$, $p<0.001$), indicating close agreement between tool set matching and MAM’s multi-judge consensus.
\section{Ablation Study}
In this section, we present additional experimental results on a subset of our dataset. By default, we sample 800 queries, with 100 queries from each category.

\noindent\textbf{Impact of In-Context Tool Demonstrations.}~Table~\ref{fewshot} shows that adding in-context demonstrations substantially benefits models with sufficient instruction-following ability. GPT-4o sees notable gains in final score, success rate, and fewer missed images, reflecting better tool calibration. Qwen2.5-VL-72B improves even more in success rate, indicating more accurate tool calls and adherence to formats. LLaMA-3.2-11B, however, shows only marginal changes, implying few-shot prompting cannot offset its core tool-planning and execution limits.

\begin{figure}[t]
    \centering
    \includegraphics[width=0.9\linewidth]{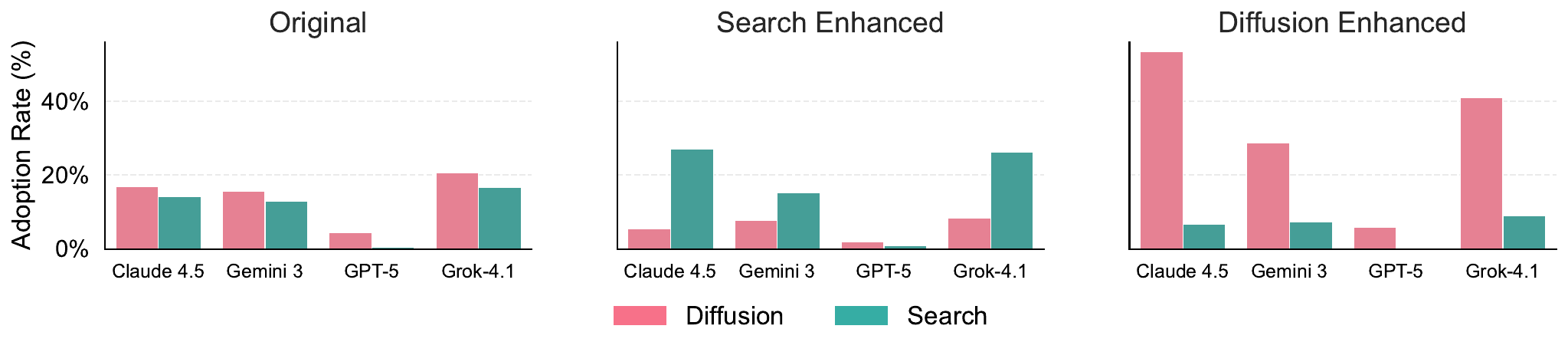}
    \caption{Impact of tool boundary on Tool Adoption Rate. Claude 4.5 and Gemini 3 denote Claude Sonnet 4.5 and Gemini 3 Pro, respectively.}
    \label{fig:boundary}
\end{figure}

\noindent\textbf{Impact of Tool Capability Boundaries.}~We identify an ambiguous region where images can be produced by either \emph{search} or \emph{diffusion}, such as “La Tour Eiffel in rain.” To quantify how models resolve this ambiguity, we evaluate three prompt variants: Original, Search Enhanced, and Diffusion Enhanced. As shown in Figure~\ref{fig:boundary}, Claude Sonnet 4.5, Grok-4.1, and Gemini 3 Pro shift toward \emph{search} under Search Enhanced prompts and strongly adopt \emph{diffusion} under Diffusion Enhanced prompts, while GPT-5 remains conservative. This indicates that tool selection is highly sensitive to capability definitions, supporting agents’ adaptation to evolving tool boundaries.

\noindent\textbf{Impact of Judge Model.}~Table~\ref{tab:judge_both} compares results under Claude Sonnet 4.5, Gemini 2.5 Pro, and GPT-5 as judges. The overall ranking remains largely consistent: top models stay strong across judges, while mid-tier and weaker systems remain lower, indicating stable cross-judge trends. This is corroborated by high inter-judge Spearman rank correlations as shown in Table~\ref{tab:judge_spearman}. Judgment differences primarily stem from score calibration rather than rank reversal. Specifically, Claude Sonnet 4.5 assigns lower scores and more missed images to weaker models; Gemini 2.5 Pro better separates systems and yields fewer missed images; and GPT-5 gives higher scores and fewer missed images to weaker models.

\begin{table}[t]
\centering
\caption{Impact of the judge model and inter-judge rank correlation. Claude, Gemini, and GPT denote Claude Sonnet 4.5, Gemini 2.5 Pro, and GPT-5, respectively.}
\label{tab:judge_both}

\begin{subtable}[t]{0.63\textwidth}
\centering
\resizebox{\textwidth}{!}{
\begin{tabular}{@{}lccc|ccc|ccc@{}}
\toprule
 & \multicolumn{3}{c|}{\textbf{Claude}}
 & \multicolumn{3}{c|}{\textbf{Gemini}}
 & \multicolumn{3}{c}{\textbf{GPT}} \\
\cmidrule(lr){2-4}\cmidrule(lr){5-7}\cmidrule(lr){8-10}
 & \textbf{FS}$\uparrow$ & \textbf{SR}$\uparrow$ & \textbf{MI}$\downarrow$
 & \textbf{FS}$\uparrow$ & \textbf{SR}$\uparrow$ & \textbf{MI}$\downarrow$
 & \textbf{FS}$\uparrow$ & \textbf{SR}$\uparrow$ & \textbf{MI}$\downarrow$ \\
\midrule
\textbf{Claude Sonnet 4} & 67.49 & \underline{84.48} & \underline{2.20} & \underline{70.24} & \underline{77.31} & 1.10 & 72.19 & \underline{89.75} & 1.49 \\
\textbf{Gemini 3 Pro} & \textbf{71.33} & \textbf{84.65} & \textbf{1.85} & \textbf{78.73} & \textbf{80.95} & \textbf{0.55} & \underline{72.42} & \textbf{89.84} & \underline{1.48} \\
\textbf{Grok-4.1} & \underline{67.69} & 81.28 & \textbf{1.85} & 68.23 & 71.55 & \underline{1.07} & \textbf{74.99} & 89.50 & \textbf{1.32} \\
\textbf{GPT-4o} & 53.06 & 69.87 & 2.65 & 60.35 & 63.05 & 1.39 & 68.37 & 83.41 & 1.50 \\
\textbf{Qwen3-VL-Plus} & 57.35 & 75.74 & 3.07 & 63.5 & 65.69 & 1.35 & 68.37 & 83.37 & 1.70 \\
\textbf{Qwen2.5-VL-72B} & 46.41 & 51.44 & 3.36 & 53.88 & 39.06 & 1.64 & 62.68 & 61.56 & 1.79 \\
\textbf{InternVL-3.5-14B} & 47.22 & 6.83 & 2.84 & 50.14 & 11.46 & 1.63 & 63.13 & 68.30 & 1.79 \\
\textbf{LLaMA-3.2-11B} & 23.69 & 25.47 & 4.12 & 29.6 & 25.35 & 2.10 & 57.03 & 66.72 & 1.86 \\
\bottomrule
\end{tabular}}
\caption{Impact of the judge model.}
\label{tab:judge_impact}
\end{subtable}\hfill
\begin{subtable}[t]{0.3\textwidth}
\centering
\resizebox{\textwidth}{!}{
\begin{tabular}{llccc}
\toprule
 &  & \textbf{Claude} & \textbf{Gemini} & \textbf{GPT} \\
\midrule
\multirow{3}{*}{\textbf{FS}} & \textbf{Claude} & --    & 0.952 & 0.970 \\
 & \textbf{Gemini} & 0.952 & --    & 0.898 \\
 & \textbf{GPT}  & 0.970 & 0.898 & --    \\
\midrule
\multirow{3}{*}{\textbf{SR}} & \textbf{Claude} & --    & 1.000 & 0.881 \\
 & \textbf{Gemini} & 1.000 & --    & 0.881 \\
 & \textbf{GPT}  & 0.881 & 0.881 & --    \\
\midrule
\multirow{3}{*}{\textbf{MI}} & \textbf{Claude} & --    & 0.922 & 0.952 \\
 & \textbf{Gemini} & 0.922 & --    & 0.946 \\
 & \textbf{GPT} & 0.952 & 0.946 & --    \\
\bottomrule
\end{tabular}

}
\caption{Spearman $\rho$ ($p<0.01$).}
\label{tab:judge_spearman}
\end{subtable}
\end{table}

\noindent\textbf{Human Agreement Study.} We sampled 400 queries from all models and asked the annotators to examine the evaluation reports of three agents, rating agreement on a 0–2 scale. The Precision Inspector, Recall Inspector, and Chief Judge achieved agreement rates of 84.00\%, 85.88\%, and 88.00\%, respectively, indicating high consistency between human and MLLM judges.

\begin{table}[htbp]
    \centering
\caption{Human evaluation on end-to-end execution. Note that fewer \textit{\# Inappr.} images do not imply a higher Success Rate, as malformed tool calls fail to generate any output.}
    \label{tab:eval_human_metrics}
    \resizebox{0.4\linewidth}{!}{
    \begin{tabular}{lccc}
        \toprule
        \textbf{Model} & \textbf{\# Inappr.} & \textbf{MI} $\downarrow$ & \textbf{FS} $\uparrow$ \\
        \midrule
        \textbf{Claude Sonnet 4.5}   & 1.02 & 0.26 & 3.60 \\
        \textbf{Claude Sonnet 4}     & 1.59 & 0.34 & 3.35 \\
        \textbf{Gemini 3 Pro}        & 0.82 & \textbf{0.06} & \textbf{3.89} \\
        \textbf{Grok-4.1}            & 1.17 & 0.10 & 3.53 \\
        \textbf{GPT-5}               & 0.78 & 0.39 & 3.63 \\
        \textbf{GPT-4o}              & 0.54 & 0.88 & 3.24 \\
        \textbf{Qwen3-VL-Plus}       & 0.70 & 0.77 & 3.01 \\
        \textbf{Qwen2.5-VL-72B}      & 0.40 & 0.48 & 2.14 \\
        \textbf{InternVL3.5-14B}     & 0.08 & 1.37 & 1.26 \\
        \textbf{LLaMA-3.2-11B}       & 0.27 & 1.16 & 1.29 \\
        \bottomrule
    \end{tabular}
    }
\end{table}

\noindent\textbf{Human Evaluation on End-to-End Execution.}~Table~\ref{tab:eval_human_metrics} reports human evaluation results on end-to-end interleaved generation. For each model, we sample 100 tool plans and execute all tool calls using implemented backends: in-context images for \emph{Reference}, nano-banana~\cite{nanobanana} for \emph{Diffusion}, Google Image Search via Serp API~\cite{serpapi} for \emph{Search}, GPT-5~\cite{gpt5} generated Python scripts for \emph{Code}, and Doubao Seedream 4.0~\cite{seedream2025seedream} for \emph{Edit}. We evaluate model performance using inappropriate images (\# Inappr.), missed images (MI), and a five-point final score (FS). Results show that Gemini 3 Pro achieves the best overall performance, followed by GPT-5 and Claude Sonnet 4.5. In contrast, InternVL3.5-14B and LLaMA-3.2-11B show low inappropriate image counts but much lower FS, mainly due to tool-call failures. MAM also strongly correlates with human evaluation, with Spearman rank correlations $\rho=0.8909$ for FS ($p<0.001$) and $\rho=0.8303$ for MI ($p<0.01$), supporting its reliability as an automatic judge.
\section{Discussion}

This work has several limitations, which we leave for future work. (1) Our interleaved generation setting focuses on text--image outputs and does not cover richer modalities such as audio or video. (2) Our toolkit is restricted to five tools and therefore does not capture broader agentic capabilities. (3) We focus on direct evaluation using MLLM-as-a-judge, and do not study alternative pipelines such as captioning images for LLM-as-a-judge.

\section{Conclusion}
In this work, we identify \emph{Agentic Tool Planning} as a key next step for interleaved generation, where MLLMs autonomously coordinate tools to produce interleaved tool plans. We introduce \name, the first benchmark that unifies hybrid image sourcing and dual query types under expert-annotated visual-critical intents. We further propose MAM, a Multi-Agent MLLM-as-a-Judge framework that disentangles tool-call precision, missed images, and response quality without requiring ground truth or end-to-end execution. Experiments on 10 state-of-the-art MLLMs show that current systems still struggle with coherent tool planning, and display notable differences in tool-use behavior.

\section*{Acknowledgements}
This work was supported by Qwen Business Unit through Alibaba Research Intern Program. We thank Alibaba Group for providing research resources and annotation support, and the annotators for their valuable assistance.

\bibliographystyle{splncs04}
\bibliography{main}

\end{document}